\documentclass[sigconf]{acmart}
\AtBeginDocument{%
  }

\setcopyright{acmlicensed}
\copyrightyear{2018}
\acmYear{2018}
\acmDOI{XXXXXXX.XXXXXXX}
\acmConference[UKAIRS]{}{September 08--09,
  2025}{Newcastle, United Kingdom}
\acmISBN{978-1-4503-XXXX-X/2018/06}

\begin{document}

\title{Embodied AI in Social Spaces: Responsible and Adaptive Robots in Complex Settings}

\author{Aleksandra Landowska}
\email{aleksandra.landowska@nottingham.ac.uk}
\orcid{0000-0001-7703-6202}
\affiliation{%
  \institution{School of Computer Science, University of Nottingham}
  \city{Nottingham}
  \country{UK}
}

\author{Aislinn D Gomez Bergin}
\email{aislinn.bergin@nottingham.ac.uk}
\affiliation{%
  \institution{School of Computer science, University of Nottingham}
  \city{Nottingham}
  \country{UK}}

\author{Ayodeji O. Abioye}
\email{ayodeji.abioye@open.ac.uk}
\orcid{0000-0003-4637-3278}
\affiliation{%
  \institution{School of Computing and Communications, The Open University}
  \city{Milton Keynes}
  \country{UK}
}
\author{Jayati Deshmukh}
\email{j.deshmukh@soton.ac.uk}
\orcid{0000-0002-1144-2635}
\affiliation{%
  \institution{School of Electronics and Computer Science, University of Southampton}
  \city{Southampton}
  \country{UK}
}
\author{Andriana Boudouraki}
\email{andriana.boudouraki@nottingham.ac.uk}
\affiliation{%
  \institution{School of Computer Science, University of Nottingham}
  \city{Nottingham}
  \country{UK}}

\author{Maria Waheed}
\email{maria.waheed@nottingham.ac.uk}
\affiliation{%
  \institution{School of Computer Science, University of Nottingham}
  \city{Nottingham}
  \country{UK}}

\author{Athina Georgara}
\email{a.georgara@soton.ac.uk}
\orcid{}
\affiliation{%
  \institution{School of Electronics and Computer Science, University of Southampton}
  \city{Southampton}
  \country{UK}
}
\author{Dominic Price}
\email{dominic.price@nottingham.ac.uk}
\orcid{0000-0001-6636-9482}
\affiliation{%
  \institution{School of Computer Science, University of Nottingham}
  \city{Nottingham}
  \country{UK}
}
\author{Tuyen Nguyen}
\email{tuyen.nguyen@soton.ac.uk}
\orcid{}
\affiliation{%
  \institution{School of Electronics and Computer Science, University of Southampton}
  \city{Southampton}
  \country{UK}
}
\author{Shuang Ao}
\email{s.ao@soton.ac.uk}
\affiliation{%
  \institution{Responsible AI, University of Southampton}
  \city{Southampton}
  \country{UK}}

\author{Lokesh Singh}
\email{lb5e23@soton.ac.uk}
\orcid{}
\affiliation{%
  \institution{School of Electronics and Computer Science, University of Southampton}
  \city{Southampton}
  \country{UK}
}
\author{Yi Dong}
\email{yi.dong@liverpool.ac.uk}
\affiliation{%
  \institution{University of Liverpool}
  \city{Liverpool}
  \country{UK}}

\author{Rafael Mestre}
\email{r.mestre@soton.ac.uk}
\affiliation{%
  \institution{School of Electronics and Computer Science, University of Southampton}
  \city{Southampton}
  \country{UK}
}
\author{Joel E. Fischer}
\email{joel.fischer@nottingham.ac.uk}
\orcid{0000-0001-8878-2454}
\affiliation{%
  \institution{School of Computer Science, University of Nottingham}
  \city{Nottingham}
  \country{UK}
}

\author{Sarvapali D. Ramchurn}
\email{sdr1@soton.ac.uk}
\orcid{0000-0001-9686-4302}
\affiliation{%
  \institution{School of Electronics and Computer Science, University of Southampton}
  \city{Southampton}
  \country{UK}
}

\renewcommand{\shortauthors}{Landowska et al.}

\begin{abstract}
  This paper introduces and overviews a multidisciplinary project aimed at developing responsible and adaptive multi-human multi-robot (MHMR) systems for complex, dynamic settings. The project integrates co-design, ethical frameworks, and multimodal sensing to create AI-driven robots that are emotionally responsive, context-aware, and aligned with the needs of diverse users. We outline the project's vision, methodology, and early outcomes, demonstrating how embodied AI can support sustainable, ethical, and human-centred futures.
\end{abstract}



\received{xxx}
\received[revised]{xxx}
\received[accepted]{xxx}

\maketitle

\section{Introduction}
As global populations continue to grow, we are facing an urgent and complex societal challenge: the widening gap between the increasing number of older adults who require care and the declining availability of human caregivers\cite{WHO2024Ageing}. While advancements in medicine and public health have enabled people to live longer, healthier lives, this demographic shift presents critical pressures on already overstretched health and social care systems. In the UK alone, projections indicate that over a million additional care staff will be needed in the coming decade\cite{HealthFoundation2024Staff}
Robots and artificial intelligence (AI) are increasingly proposed as part of the solution\cite{BEIS2021AI}. Their potential to assist with repetitive tasks, provide companionship, and support monitoring in care environments has been widely explored in both academic research and industry\cite{Servaty2020RoboticsNursing, Papadopoulos2020SociallyAssistive, Guemghar2022MentalHealthRobots,Koh2021SocialRobotsDementia,Warmbein2023MobilizationRobots}. However, despite promising prototypes and pilot studies, most robotic systems have failed to achieve meaningful integration into care settings. One key reason for this lack of success lies in a narrow focus on technical performance, often overlooking the broader socio-technical, ethical, and emotional complexities of caregiving\cite{Greenhalgh2017Framework}.

This project proposes a novel approach: the development of responsible and adaptive multi-human multi-robot (MHMR) systems designed specifically for complex social environments such as care homes. Our aim is to move beyond automation for efficiency, and towards building robotic systems that are context-aware, emotionally empathetic, socially intelligent, ethically grounded, and capable of adapting to individual needs, emotions, mental states and preferences, supporting both residents and caregivers in a collaborative manner.

By embedding responsible innovation principles from the beginning—considering ethics, acceptability, regulatory compliance, and user empowerment—we seek to address the socio-technical barriers that have hindered previous efforts. Our approach combines co-design with stakeholders, real-world scenario prototyping, physiological and behavioural sensing, and explainable adaptive AI. We argue that the successful adoption of robotics in care settings requires more than just functional systems; it requires trusted, transparent, empathetic and human-centred technologies that align with the values and realities of the people they are meant to support.

This paper outlines our vision, objectives, and predicted outcomes in designing such systems. We highlight the interdisciplinary methodology underpinning the project and provide a roadmap for designing, validating, and deploying embodied AI in socially sensitive environments. Our goal is to ensure that robotics not only alleviate burden but do so in a way that respects human dignity, enhances relationships, and contributes to sustainable care futures.
\section{Methodology}
This is a multidisciplinary project that brings together expertise in human-computer interaction, robotics, cognitive neuroscience, AI, ethics and healthcare. The overall aim of the project is to explore how MHMR adaptive systems can be responsibly integrated into complex social environments, particularly within health and social care.

To achieve this, the project has been structured into five work packages (WPs), with co-creation embedded throughout the whole project stage (see Figure 1 for the overview). We work closely with diverse stakeholders—including older adults, carers, healthcare professionals, care home residents as well as technologists, designers and ethicists—to ensure the technologies we develop are informed by real-world experiences, needs, and values.

\begin{figure}
     \centering
     \includegraphics[width=1.0\linewidth]{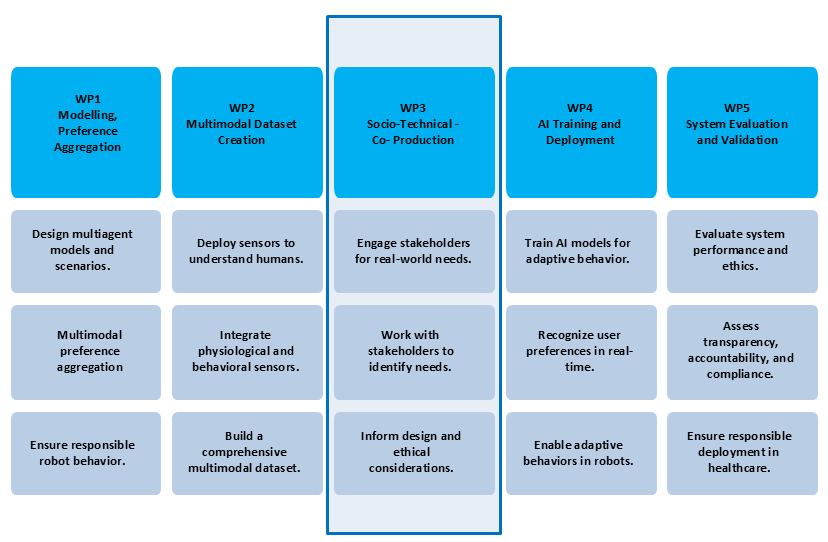}
    \caption{Overview of all Work Packages in the project. The diagram summarises the five Work Packages, including their titles and key objectives. Work Package 3 (Socio-Technical Co-Production and Implementation) is highlighted at the centre, reflecting its cross-cutting role. It informs and supports all other Work Packages through continuous stakeholder engagement across all stages of the project.
    }
     \label{fig:fnirs-contrast}
\end{figure}

\section{Discussion}
\subsection{Impact}
This project addresses the increasing demand for care in ageing societies and the shortage of available human caregivers. By developing multimodal multi-agent robotic systems that can understand and adapt to human needs, we aim to enhance the quality and efficiency of care without replacing essential human interaction. Our approach ensures that robots support rather than burden care environments, contributing to a more sustainable care infrastructure. 
\subsection{Outcomes}
The project will produce several key outputs:
 \\ - Adaptive robotic systems trained using multimodal sensor data
 \\ - A multimodal dataset that integrates neurophysiological and behavioural data
 \\ - AI models that interpret user preferences, emotional states, and care-related requirements
 \\ - Realistic care scenarios and evaluation protocols
 \\ - Ethical frameworks and practical guidance for responsible deployment
 \\ - Demonstrator showcasing an adaptive and responsible multi-agent system capable of adjusting its behaviour in response to users’ cognitive and emotional states
 \\ - Tools to help care providers and policymakers assess and adopt responsible robotic solutions
\subsection{Significance}
This research advances human-robot interaction by combining adaptive technology with ethical and social responsibility. It offers a practical model for deploying AI in care settings and other sensitive areas, with a focus on ethics and real-world relevance. The project has the potential to shape the future of healthcare by demonstrating how adaptive robotic systems can improve the quality of life for residents, ease the workload of carers, and provide reassurance and support for families. It also contributes to the broader development of responsible adaptive robotics across domains, supporting the wellbeing of individuals, communities, and societies.

\section{Acknowledgments}
This work was supported by the Engineering and Physical Sciences Research Council [grant number EP/Y009800/1], through funding from Responsible Ai UK (KP0016).

\bibliographystyle{ACM-Reference-Format}
\bibliography{paper}


\end{document}